\documentclass{article}
\usepackage[final]{fixme}

\usepackage[preprint]{nips_2018}

\usepackage[utf8]{inputenc} 
\usepackage[T1]{fontenc}    
\usepackage{hyperref}       
\usepackage{url}            
\usepackage{booktabs}       
\usepackage{amsfonts}       
\usepackage{nicefrac}       
\usepackage{microtype}      

\usepackage{microtype}
\usepackage{graphicx}
\usepackage{booktabs} 
\usepackage{threeparttable} 
\usepackage{amsmath}
\usepackage{amssymb}	
\usepackage{siunitx}
\usepackage{subcaption}

\title{Neural Message Passing with Edge Updates for Predicting Properties of Molecules and Materials}

\author{
  Peter Bjørn Jørgensen\\
  Department of Applied Mathematics and Computer Science\\
  Technical University of Denmark\\
  \texttt{pbjo@dtu.dk} \\
  \And
  Karsten Wedel Jacobsen\\
  Department of Physics\\
  Technical University of Denmark\\
  \texttt{kwj@fysik.dtu.dk} \\
  \And
  Mikkel N. Schmidt\\
  Department of Applied Mathematics and Computer Science\\
  Technical University of Denmark\\
  \texttt{mnsc@dtu.dk} \\
}

\makeatletter
 \begingroup
 \catcode`\_=\active
 \protected\gdef_{\@ifnextchar|\subtextup\sb}
 \endgroup
 \def\subtextup|#1|{\sb{\textup{#1}}}
 \AtBeginDocument{\catcode`\_=12 \mathcode`\_=32768 }
\makeatother

\newcommand{\figref}[1]{\figurename~\ref{#1}}
\begin{document}

\maketitle

\begin{abstract}
Neural message passing on molecular graphs is
one of the most promising methods for predicting formation energy and other properties of
molecules and materials.
In this work we extend the neural message passing model with an edge update network which allows the information exchanged between atoms to depend on the hidden state of the receiving atom.
We benchmark the proposed model on three publicly available datasets (QM9, The Materials Project and OQMD) and show that the proposed model yields superior prediction of formation energies and other properties on all three datasets in comparison with the best published results.
Furthermore we investigate different methods for constructing the graph used to represent crystalline structures and we find that using a graph based on K-nearest neighbors achieves better prediction accuracy than using maximum distance cutoff or the Voronoi tessellation graph.
\end{abstract}

\section{Introduction}

The current workhorse for screening of new molecules and materials is Density Functional Theory (DFT),
but machine learning methods show a potential for trading DFT accuracy for a several orders of magnitude decrease in computation time.
Until recently machine learning methods for predicting properties of molecules and materials have been based
on hand crafted feature descriptors, such as Coulomb matrix \citep{Rupp2012-rq}, bag-of-bonds \citep{Hansen2015-rp}, fingerprints (e.g. \citealt{Rogers2010-ua})  and histogram of angles \citep{Faber2017-ur}. With the availability of large databases of molecules and materials we are now seeing a shift towards data-driven representation learning as we have seen in the computer vision field.

The graph neural network model was introduced by \citet{Gori2005-dt,Scarselli2009-wl} and regained attention when \citet{Li2015-ly} expanded the model for different graph problems. In the last few years a number of graph-based models for molecules have been proposed \citep{Duvenaud2015-vd, Kearnes2016-te, Schutt2017-hk, Schutt2017-af, Gilmer2017-wp}.
These models can all be cast into the framework of message passing on a molecular graph as shown by \citet{Gilmer2017-wp}.

The molecular graphs used as input for these models are either the topological graph defined by the chemical bonds (with or without bond lengths) or a fully connected graph where the pairwise distances between all the atoms are used as edge features.
The topological approach does not directly apply to crystalline structures because the chemical bonds are less well-defined and the fully connected approach is impossible because the structure is infinite.
The first application of neural message passing for materials is SchNet \citep{Schutt2017-kq} where a constant cutoff distance is used. As noted by \citet{Schutt2017-kq} this may lead to ``isolated atoms'' if the cutoff distance is too small and the computational burden increases with the size of the cutoff. This motivates us to investigate different ways to define the molecular graphs based on nearest neighbor cutoff. Concurrently with the work presented in this paper \citet{Xie2018-hz} have also employed neural message passing on graphs based on Voronoi tessellation and K-nearest neighbors, but they do not include a comparison between the methods.

The edge neural network with set-to-set readout function (enn-s2s) \citep{Gilmer2017-wp} demonstrated state of the art prediction accuracy on the 13 properties of the QM9 \citep{Ramakrishnan2014-ey, Ruddigkeit2012-dc} dataset consisting of 134k molecules. The recently proposed SchNet \citep{Schutt2017-af, Schutt2017-kq} network improves the accuracy on eight out of the twelve properties.
In both of these models the information exchanged between the atoms in the message passing scheme depends on the representation of the sending atom and the edge feature on which the message is passed, but is independent of the representation of the receiving atom.
\citet{Gilmer2017-wp} proposed the use of a ``pair message'' network that includes the state of the receiving atom, but its predicting performance is inferior to the enn-s2s network.
In this work we propose to extend the SchNet model with an edge update network such that the edge feature depends on the representation of the atoms that the edge connects. This in turn means the information exchanged between the atoms also depends on the receiving atom.
Edge update networks was also utilised in the Weave module proposed by \citet{Kearnes2016-te}, but there the edges are forced to be undirected and the prediction accuracy is below that of enn-s2s \citep{Gilmer2017-wp}.

We benchmark the proposed edge update network on QM9 molecules, Materials Project \citep{Jain2013} and OQMD \citep{Saal2013-mi, Kirklin2015-xm} datasets and it shows an improvement over current state of the art results on all three datasets.
For crystalline structures we show that it is beneficial to use a K-nearest neighbor graph rather than a graph defined by a constant cutoff distance as used in previous work \citep{Schutt2017-kq}.

The paper is organised as follows. We present the proposed model within the neural message passing framework in Section~\ref{sec:mpnn} and we show how it is different from other models within the same framework in Section~\ref{sec:related}. We introduce the three datasets in Section~\ref{sec:datasets}, which are used for benchmarking the model in Section~\ref{sec:results} and we conclude the paper in Section~\ref{sec:conclusion}.

\section{Message Passing Neural Networks}\label{sec:mpnn}
We describe message passing neural networks similarly to \citep{Gilmer2017-wp} as a model operating on a graph $G$ with vertex features $x_v$ and edge features $\varepsilon_{vw}$. Each vertex has a hidden state $h_v^t$ and each edge has a hidden state $e_{vw}^t$ which are updated in a number of interaction steps $T$. Vertices are updated using a message function $M_t(\cdot)$ and a state transition function $S_t(\cdot)$
\begin{align}
	m_{v}^{t+1} &= \sum_{w \in N(v)} M_{t}(h_v^t, h_w^t, e_{vw}^t), \label{eq:message_func}\\
	h_v^{t+1} &= S_t \left( h_v^t, m_{v}^{t+1} \right),
	\label{eq:updates}
\end{align}
where $N(v)$ denotes the neighbourhood of $v$, i.e. the vertices that have an edge to $v$.
The edges are updated by an edge update function $E_t(\cdot)$ that depends of the previous edge state and the states of the sending and receiving vertices of the current step
\begin{equation}
	e_{vw}^{t+1} = E_t\left( h_v^{t+1}, h_w^{t+1}, e_{vw}^{t} \right).
	\label{eq:edge_update}
\end{equation}
After the $T$ interaction steps a readout function $R(\cdot)$ is applied which maps the set of vertex states into a single entity
\begin{equation}
	\hat y = R\left(\{h^T_v \in G \}\right).
	\label{eq:readout}
\end{equation}
The readout function must be invariant to permutation of the vertex set, which is often achieved via summation over the vertex features. The functions $M_t$, $S_t$, $E_t$ and $R$ are all implemented as neural networks with trainable parameters and can be optimised using gradient descent. Some models, for example Deep Tensor Neural Network \citep{Schutt2017-hk}, use weight sharing across the interaction steps, thus $M_t \equiv M$, $S_t \equiv S$ and $E_t \equiv E$ for all time steps $t$.
A range of graph convolution models \citep{Duvenaud2015-vd, Li2015-ly, Battaglia2016-mj, Kearnes2016-te, Schutt2017-hk} including Laplacian based models \citep{Bruna2013-dv} can be cast into this message passing framework as shown by \citet{Gilmer2017-wp}.

\subsection{SchNet with Edge Update Network}

We now describe how our model fits into the message passing framework described above.
As vertex input features, $x_v$, we use the atomic numbers which are translated into an embedding vector for each atomic number as in \citep{Schutt2017-hk, Schutt2017-af, Schutt2017-kq}. The initial hidden state $h_v^0=\ell(x_v)$ is thus the result of a lookup function $\ell:\mathbb{Z}\to \mathbb{R}^{C}$.
The hidden state is a representation of the atom and its chemical environment and the idea behind the message passing interaction steps is to refine this representation based on the surrounding atoms and their chemical environments.
Since the interaction between atoms depends on the distance between them (Coulomb's law) we use the interatomic distances as initial edge features.
We then expand the initial edge feature using a radial basis function.
Denoting the distance between atom $v$ and $w$ as $d_{vw}$, the edge between node $v$ and $w$ in the graph has initial feature vector
\begin{align}
	(\varepsilon_{vw})_{k} = \exp \left(-\frac{(d_{vw} - (-\mu_|min|+ k \Delta))^2}{\Delta} \right),k=0\ldots k_|max|
	\label{eq:r}
\end{align}
where $\mu_|min|$, $\Delta$, and $k_|max|$ are chosen such that the centers of the functions covers the range of the input features. In all our experiments we set $\mu_|min|=\SI{0}{\angstrom}, \Delta=\SI{0.1}{\angstrom}, k_|max|=150$. This expansion makes it easier for the neural network model to decorrelate input and output similar to how 1-hot-encoding is preferred over integer coding for categorical features.

The role of the hidden edge representation $e_{vw}^{t}$ is to control how the two connected atoms interact.
The idea of using an edge update network is to let the updated atomic hidden states influence this interaction.
We use an edge update function at each interaction step implemented as a two layer feed-forward neural network. The input to the network is the concatenation of the edge representation and the hidden states of the receiving and sending vertices. Thus
\begin{align}
	e_{vw}^0 &= E_0(h_v^0,h_w^0, \varepsilon_{vw}) = g(W_|E2|^0 g(W_|E1|^0(h_v^0; h_w^0; \varepsilon_{vw}))),
\end{align}
and similarly for the subsequent steps
\begin{align}
	e_{vw}^{t+1} = g(W_|E2|^{t+1} g(W_|E1|^{t+1}(h_v^{t+1}; h_w^{t+1}; e_{vw}^t))),
\end{align}
where $g(x)=\ln( \operatorname{e}^x+1)- \ln(2)$ is the shifted soft-plus activation function $(\cdot;\cdot)$ denotes vector concatenation and $\{W_|E1|^{0}, W_|E2|^{0},W_|E1|^{t+1}, W_|E2|^{t+1}\}$ are trainable weight matrices. This update makes the network edges directional.

The message function itself is only a function of the sending node and can be written as
\begin{align}
	M_{t}(h_v^t, h_w^t, e_{vw}^t) &= M_{t}(h_w^t, e_{vw}^t) =  (W_{1}^{t} h_w^{t}) \odot g(W_{3}^{t}g(W_{2}^{t}e_{vw}^t)),
	\label{eq:msg_function}
\end{align}
where $\odot$ denotes element-wise multiplication and $\left\{ W_1^t, W_2^t, W_3^t \right\}$ are trainable weight matrices.
This message function enables the interpretation of the function on the right hand side of the element-wise multiplication as a filter-generating function \citep{Schutt2017-af, Schutt2017-kq} $f_g^t(e_{vw}^t)=g(W_{3}^{t}g(W_{2}^{t}e_{vw}^t))$, a continuous analogous to the filters applied over a discrete domain in a convolutional neural network for image processing.
We visualize this function towards the end of this section.
The shifted soft-plus activation function is chosen to follow \citet{Schutt2017-af, Schutt2017-kq} and can be seen as an infinitely differentiable alternative to the rectified linear unit (ReLU) activation function.

\begin{figure}[tp]
	\centering\includegraphics[width=1.0\textwidth]{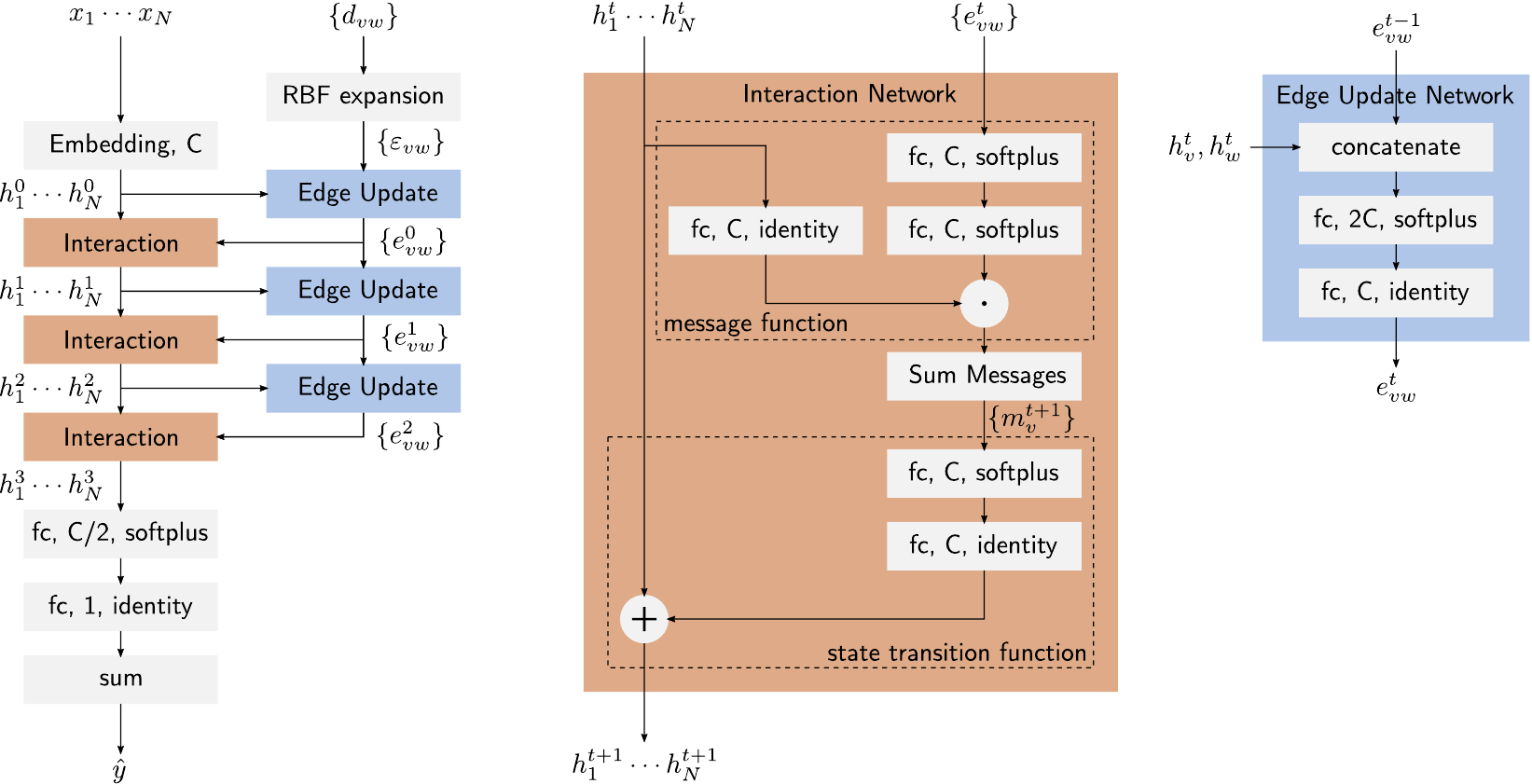}
	\caption{Flow of computations of the proposed model. The dimension of the hidden state for each atom state is $C$ and the ``fc''-blocks are fully connected layers annotated with the output dimension and the applied activation function.}
	\label{fig:flowchart}
\end{figure}

The state transition function applies a two layer neural network on the sum of incoming messages and adds this to the current hidden state as in Residual Networks \citep{He2015-on}:
\begin{equation}
	S_t \left( h_v^t, m_{v}^{t+1} \right) = h_v^t + W_5^t g( W_4^t m_{v}^{t+1}),
	\label{eq:state_transition}
\end{equation}
where $\left\{ W_4^t, W_5^t\right\}$ are trainable weight matrices.
After a number of interaction steps $T$ all the information about the property we want to predict must be contained in the set of hidden node states. We apply a readout function for which we use a two layer neural network that maps the hidden representation to a scalar and finally we sum over the contribution from each atom, i.e.
\begin{equation}
	R\left( \left\{ h_v^T \in G \right\} \right) = \sum_{h_v^T \in G} W_{7} g(W_{6} h_v^T),
	\label{eq:readout_func}
\end{equation}
where $\left\{ W_7, W_6\right\}$ are also trainable weight matrices.
The model architecture and flow of computations is illustrated in \figref{fig:flowchart}.
The original SchNet architecture is obtained if we ``remove'' the edge update function, i.e. by setting $E_t( h_v^t, h_w^t, e_{vw}^{t-1})= e_{vw}^{ t-1} $. This property enables direct comparison between the two models.

\begin{figure}[tp]
	\begin{minipage}[t]{0.48\textwidth}
	\centering\includegraphics[width=1.0\textwidth]{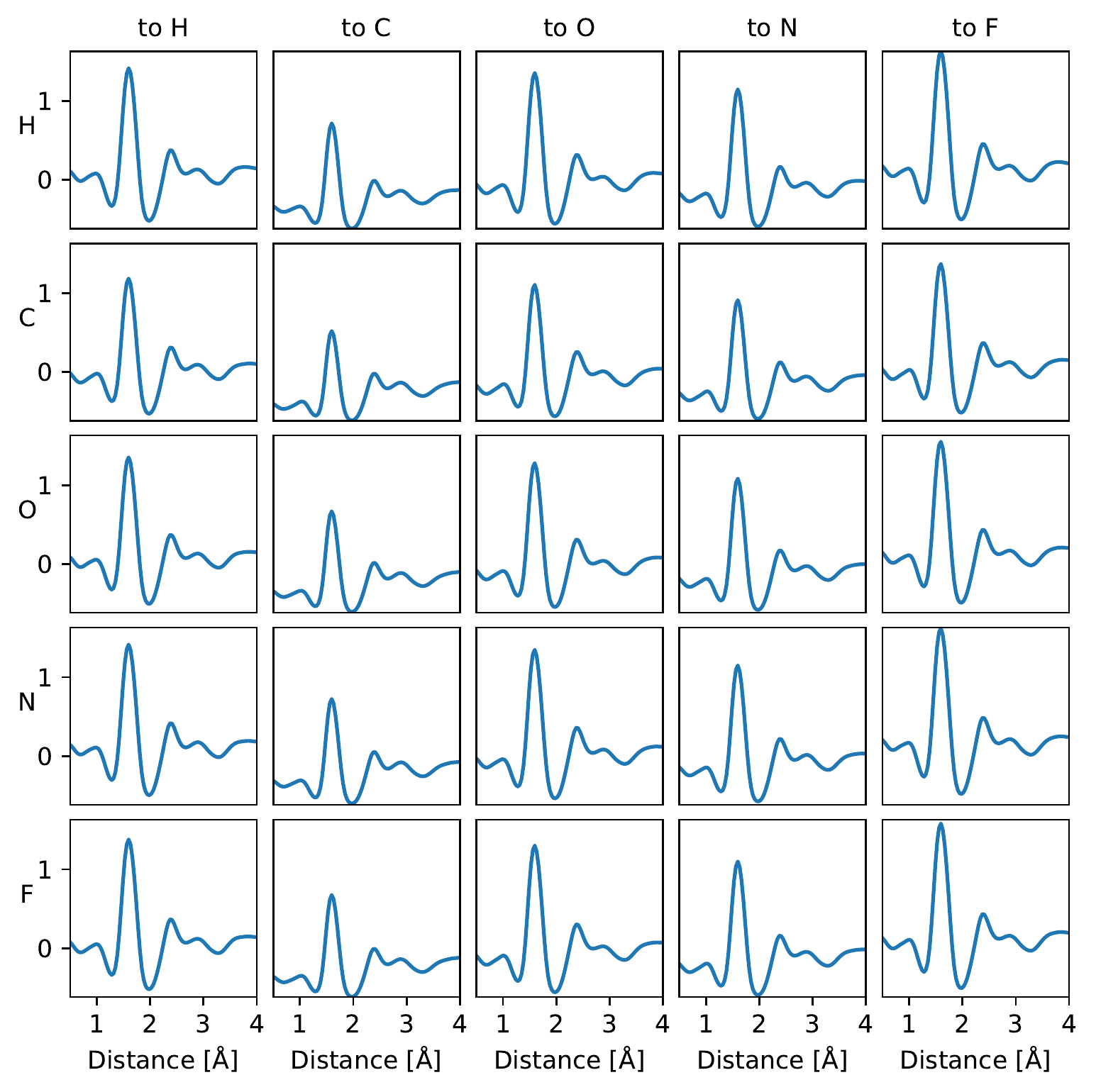}
	\captionof{figure}{An example of one of the learned filters (out of 64) at the first layer of the message passing architecture trained on prediction of formation energy $U_0$ of the QM9 dataset. The filter depends on the embedding of the sending (the rows) and the receiving (the columns) nodes.}
	\label{fig:filters}
	\end{minipage}\hfill
	\begin{minipage}[t]{0.48\textwidth}
	\centering\includegraphics[width=1.0\textwidth]{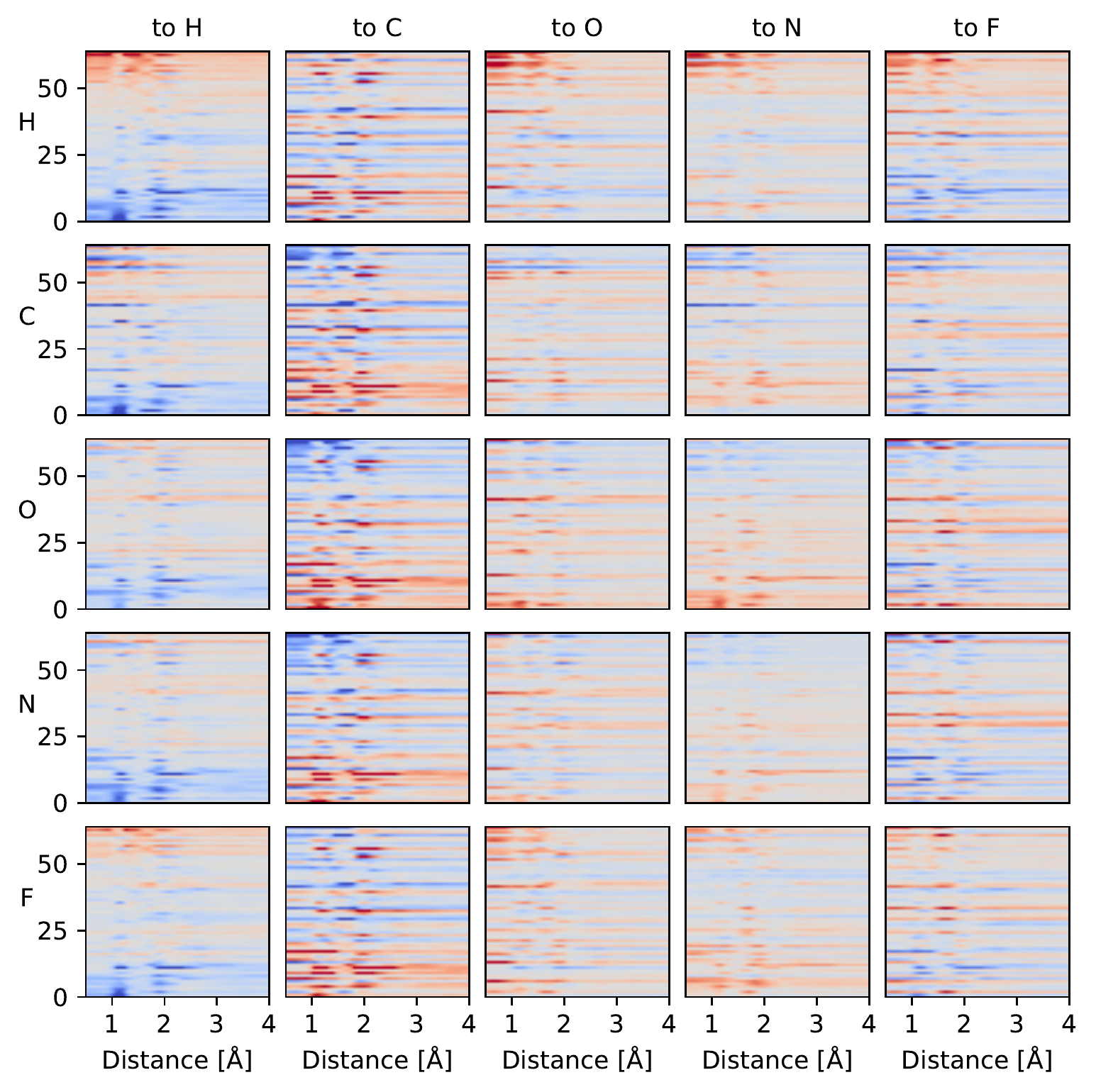}
	\captionof{figure}{Filter variation for all 64 learned filters at the first layer of the message passing architecture trained on prediction of formation energy $U_0$ of the QM9 dataset. Each row of pixels corresponds to one of the 64 filters and the color encodes the deviation (\num{-0.4} is blue and \num{0.4} is red) from the corresponding average filter (averaged over sending and receiving atom). The enumeration of the filters is arbitrary and we have sorted them according to the value of the deviation at 1.0 for the filter H$\to$H. The filter generating function shows a higher dependence on the species of the receiving atom than the species of the sending atom.}
	\label{fig:all_filters}
	\end{minipage}
\end{figure}

We want to visualize the filter generating function to qualitatively assess what the model learns, but because of the edge update network, the edge feature of the first layer $e_{vw}^0$ does not only depend on the distance $d_{vw}$ between the two atoms, but also on the atom embeddings, i.e.
\begin{align}
	f_g^0(d_{vw}) = f_g^0(E_0(h_v^0,h_w^0, \operatorname{RBF}(d_{vw}))),
	\label{eq:fg_function}
\end{align}
where $\operatorname{RBF}(\cdot)$ is the radial basis function expansion defined in \eqref{eq:r}.
We can now plot the filter response of the first layer as a function of the sending and receiving atoms and the distance between them as shown in \figref{fig:filters} and \figref{fig:all_filters}. We see that the filter-generating function is almost identical across the sending atom species but the variation across the receiving atoms is more significant. This indicates that the model has learned to shape the messages depending on the receiving atom. The ability to condition the filter on the pair of sending and receiving atoms is the key difference between the proposed model and the SchNet model which is using the same filter for all combinations of sending and receiving atoms.

\section{Related Work}\label{sec:related}
Our model is closely related to the other message passing neural network models used for molecular properties prediction.
However, the only message passing neural network that has explored the use of edge updates is the Weave module \citep{Kearnes2016-te}.
The message function is $M_{t}(h_v^t, h_w^t, e_{vw}^t)=\alpha (W_1^t e_{vw}^t)$ and the state transition function is $S_t(h_v^t, m_v^{t+1}) = \alpha ( W_3^t ( \alpha (W_2^t h_v^t); m_v^{t+1}) )$ where $\alpha(\cdot)$ is the rectified linear unit activation function, $(\cdot;\cdot)$ denotes vector concatenation and $\left\{ W_1^t, W_2^t, W_3^t \right\}$ are weight matrices.
Unlike our model where the edge updates are interleaved with the node updates, the edge updates are done in parallel, i.e.
\begin{align}
	e_{vw}^{t+1} &= E_t\left( h_v^t, h_w^t, e_{vw}^{t} \right) \\
	&= \alpha(W_6^t(\alpha(W_4^t(h_v,h_w))+\alpha(W_4^t(h_w,h_v)), \alpha(W_5^t e_{vw}^t))) \nonumber
\end{align}
Another difference is that the edge updates in the Weave module are invariant to permutation of $h_v$ and $h_w$ and the edges are thus undirected by design.
The Weave module does not use the Euclidean distance between the atoms as input features, but uses the graph distance on the chemical graph. The model was tested on a range of different classification tasks as well as drug efficacy, photovoltaic efficiency and solubility regression tasks.
The model was reimplemented by \citet{Faber2017-ur} and \citet{Gilmer2017-wp}. The version by \citet{Faber2017-ur} includes a few modifications. The edges are no longer forced to be undirected, i.e.
\begin{align}
	e_{vw}^{t+1} &= E_t\left( h_v^t, h_w^t, e_{vw}^{t} \right) \\
	&= \alpha(W_6^t(\alpha(W_4^t(h_v,h_w)), \alpha(W_5^t e_{vw}^t))) \nonumber
\end{align}
and the Euclidean distance between the atoms is included in the message function by division of the edge feature with a range of different powers of the distance $d_{vw}$:
\begin{align}
	M_{t}(h_v^t, h_w^t, e_{vw}^t)=\alpha \left( \underset{k \in \left\{ 0,1,2,3,6 \right\}}{\operatorname{concatenate}} \left( \frac{W_1^te_{vw}^t}{d_{vw}^k} \right) \right)
\end{align}
Both versions of the graph convolution model based on the Weave module shows similar performance on the QM9 regression benchmark \citep{Gilmer2017-wp, Faber2017-ur}, but the mean absolute prediction error is significantly higher than that of the enn-s2s and SchNet models.

\section{Datasets}\label{sec:datasets}
We benchmark the proposed model on three publicly available datasets.
\paragraph{Quantum Machines 9 (QM9) \citep{Ramakrishnan2014-ey, Ruddigkeit2012-dc}}  The dataset contains \num{133885} examples of stable small organic molecules with up to 7 heavy atoms (CONF) and up to 29 atoms in total including H.
The 12 target properties for each example are shown in Table \ref{tab:qm9-properties}. All properties are calculated at the B3LYP/6-31G(2df,p) level of quantum chemistry.
Following \citet{Gilmer2017-wp, Schutt2017-kq} we randomly select a training set of \num{110000} molecules, a validation set of \num{10000} molecules and the remaining \num{13885} examples are used for testing. The validation set is used for early stopping and model selection.
\paragraph{The Materials Project \citep{Jain2013}} This dataset contains geometries and formation energies of \num{69640} inorganic compounds with input structures taken from the ICSD database \citep{Bergerhoff1983-me}. We use the latest version of the database (version 2.0.0). The target property is heat of formation. The number of examples is reduced to \num{69539} after we exclude all materials with noble gases (He, Ne, Ar, Kr, Xe) because they occur so infrequently in the dataset. This brings the number of different elements in the dataset down to 84. Following \citep{Schutt2017-kq} we randomly sample \num{60000} of the examples to use as the training set. Of the remaining examples we use a set of \num{5000} for validation and the remaining \num{4539} examples are used for testing.
\paragraph{Open Quantum Materials Database (OQMD) \citep{Saal2013-mi, Kirklin2015-xm}} Is also a database of inorganic structures. We have extracted the database from the supplemental material of \citep{Ward2017-ff}. Again we consider materials with noble gases as outliers and we also exclude (highly unstable) materials with a heat of formation of more than \SI{5}{\electronvolt \per atom}. We thus exclude \num{210} out of \num{435792} examples. As with the materials project we use 20\% (\num{87116}) of the examples as a test set, \num{5000} for validation and the remaining \num{343466} for training.
\begin{table}
	\centering
	\caption{Target properties of the QM9 dataset.}
	\label{tab:qm9-properties}
	\begin{tabular}{ll}
		\toprule
		Target & Description 			\\
		\midrule
		$\varepsilon_|HOMO|$  	&Energy of highest occupied molecular orbital (HOMO)  \\
		$\varepsilon_|LUMO|$	&Energy of lowest occupied molecular orbital (LUMO)  \\
		$\Delta \varepsilon$	&Difference between LUMO and HOMO  \\
		ZPVE					&Zero point vibrational energy  \\
		$\mu$					&Dipole moment 	\\
		$\alpha$				&Isotropic polarizability 	\\
		$\langle R^2 \rangle$	&Electronic spatial extent 	\\
		$U_0$					&Internal energy at \SI{0}{K}  \\
		$U$						&Internal energy at \SI{298.15}{K}  \\
		$H$						&Enthalpy at \SI{298.15}{K}  \\
		$G$						&Free energy at \SI{298.15}{K}  \\
		$C_v$					&Heat capacity at \SI{298.15}{K} 	\\
	\end{tabular}
\end{table}

\section{Results}\label{sec:results}
We train the models with ADAM \citep{Kingma2014-tr} with initial learning rate \num{5e-4} or \num{1e-4} (when the higher learning rate leads to instability) and decrease the learning rate by multiplying with $0.96$ every \num{100000} gradient steps. We use a minibatch size of 32, run the optimization for up to \num{1e7} gradient steps, compute the validation error every \num{50000} steps and terminate if there was no improvement within the last \num{1e6} steps.
When applying the model on the two materials datasets (Materials Project and OQMD) we use an average over messages rather than a sum in \eqref{eq:message_func} because \citet{Schutt2017-kq} found that it increases the stability of the optimization. For these datasets (as well as for some of the QM9 properties, specifically $\varepsilon_|HOMO|$, $\varepsilon_|LUMO|$, $\Delta \varepsilon$, ZPVE, $\mu$, $\alpha$ and $\langle R^2 \rangle$) we also take the average rather than the sum in the readout function \eqref{eq:readout_func} because the target property is formation energy {\em per atom}. When the readout function uses a sum we first estimate the target mean and standard deviation atom-wise, i.e. $\hat{\mu} = \sum_i \frac{t_i}{n_i}, \hat{\sigma}=\sqrt{\sum_i(\frac{t_i}{n_i}-\hat{\mu})^2}$ where $t_i$ and $n_i$ are the target and number of atoms for the $i$th training sample. The targets are then normalised using $\tilde{t_i}=\frac{t_i-n_i \hat{\mu}}{\hat{\sigma}}$ such that each scalar of the sum \eqref{eq:readout_func} is expected to have zero mean and unit variance for a given sample with known $n_i$. When the readout function uses the average rather than a sum the targets are normalised to zero mean and unit variance in the ``usual'' way, irrespectively of the number of atoms. For the QM9 experiments we use a hidden state representation of dimension $C=64$ and for the two materials datasets we increase the dimensionality to $C=256$ because the number of different elements in these datasets is 84 rather than 5 in QM9. In agreement with the results shown in \citep{Schutt2017-kq} and also noticed in \citep{Gilmer2017-wp} we did not see a gain in prediction accuracy when using more than 3 interaction steps, so the results presented here are are all with $T=3$.

\subsection{Edge Update Network}
To assess the effect of the edge update network we train the proposed model with and without the edge update network to predict formation energies of the three datasets. Without the edge update network the model reduces to the SchNet model. For the two materials datasets (Materials Project and OQMD) we use a cutoff distance of \SI{5}{\angstrom} as used in \citep{Schutt2017-kq} when constructing the graphs.
In the benchmark we also include a state of the art non-deep learning, graph-based method, which creates a graph based on the Voronoi diagram, extracts a number of features from the graph and uses a random forest regression model for making predictions and we refer to this method as V-RF \citep{Ward2017-ff}. The model also uses the spatial information, but the interatomic distances are normalised such that the model's predictions are independent of the scaling. We benchmark all three models on the same training and test set, but V-RF also uses the validation set for training because early stopping is not used for this model.
The mean absolute error for the test set predictions are shown in Table \ref{tab:vanilla_vs_edge}.
In all three benchmarks we observe a big improvement in prediction accuracy when using the edge update network. This is not only due to the increase in number of parameters, because merely increasing the number of interaction steps does not improve the performance of the model.

We also train the proposed model to predict the 12 properties of the QM9 dataset. The results of SchNet \citep{Schutt2017-kq} and enn-s2s \citep{Gilmer2017-wp} are included as references.
The results are shown in Table \ref{tab:qm9_properties}.
The proposed model demonstrates a significant improvement in the prediction of 9 out of 12 properties and matches the existing results for the remaining 3.

\begin{table}
	\centering
	\caption{Mean absolute error of formation energy predictions for V-RF, SchNet and the proposed model. For QM9 the error is in \si{meV} and for the Materials Project and OQMD the numbers are in \si{meV/atom}. The lowest error is highlighted in bold. We have obtained the V-RF results by running the implementation provided by the authors \citep{Ward2017-ff}, while SchNet results have been obtained by running our own SchNet implementation. The numbers in parenthesis are the estimated 95th percentile, which have been obtained by sampling the test set (with replacement) \num{1e6} times.}
	\label{tab:vanilla_vs_edge}
	\begin{tabular}{lllll}
		\toprule
		Model & QM9 $U_0$& Mat.Proj.  &OQMD\\
		\midrule
		V-RF & - & 76.8 (79.8) & 74.5 (75.1) \\
		SchNet & 13.6 (14.2) & 31.8 (33.3) & 27.5 (27.9) \\
		Proposed Model &  \textbf{10.5} (11.1) & \textbf{22.7} (24.0) & \textbf{14.9} (15.2) \\
		\bottomrule
	\end{tabular}
\end{table}
\begin{table}
	\centering
	\caption{Mean absolute error of predictions for different target properties of the QM9 dataset using 110k training examples. The lowest error is highlighted in bold. SchNet and enn-s2s results are from \citep{Schutt2017-kq} and \citep{Gilmer2017-wp} respectively. The numbers in parenthesis are the estimated 95th percentile, which have been obtained by sampling the test set (with replacement) \num{1e6} times.}
	\label{tab:qm9_properties}
	\begin{tabular}{llllll}
		\toprule
		Target & Unit & SchNet & enn-s2s & Proposed & (95th) \\
		\midrule
		$\varepsilon_|HOMO|$  	& \si{meV}				& 41	& 43	& \textbf{36.7} & (37.3)\\
		$\varepsilon_|LUMO|$	& \si{meV}				& 34	& 37	& \textbf{30.8} & (31.3)\\
		$\Delta \varepsilon$	& \si{meV}				& 63	& 69	& \textbf{58.0} & (58.9)\\
		ZPVE					& \si{meV}				& 1.7	& \textbf{1.5}	& \textbf{1.49} 			& (1.52)\\
		$\mu$					& \si{Debye}			& 0.033	& 0.030	& \textbf{0.029}&  (0.029)\\
		$\alpha$				& \si{Bohr^3}			& 0.235	& 0.092	& \textbf{0.077}&  (0.082)\\
		$\langle R^2 \rangle$	& \si{Bohr^2}			& \textbf{0.073}	& 0.180	& \textbf{0.072} 		& (0.075)\\
		$U_0$					& \si{meV}				& 14	& 19	& \textbf{10.5} & (11.1)\\
		$U$						& \si{meV}				& 19	& 19	& \textbf{10.6} & (11.2)\\
		$H$						& \si{meV}				& 14	& 17	& \textbf{11.3} & (11.9)\\
		$G$						& \si{meV}				& 14	& 19	& \textbf{12.2} & (12.7)\\
		$C_v$					& \si{cal\per mol  K}	& \textbf{0.033}	& 0.040	& \textbf{0.032} 		& (0.033)\\
		\bottomrule
	\end{tabular}
\end{table}

\subsection{Choosing The Cutoff}
We want to investigate the importance of choosing a cutoff when constructing the graph used as input to the neural message passing models. The choice of cutoff is important because the computational complexity of the algorithm increases linearly with the number of edges in the graph. On the other hand if the number of edges is too small the interaction between the nodes of the graph may be too limited. In \citep{Schutt2017-kq} and in our experiments above a constant cutoff distance of \SI{5}{\angstrom} is used.
We use the OQMD dataset for this experiment and use the formation energy as the target for predictions.
With this cutoff distance some of the atoms of the dataset becomes isolated and further increasing the cutoff distance comes with prohibitive computational cost. Alternatively we can use a K-nearest cutoff method such that each atom receives messages from the K nearest neighboring atoms no matter how far away they are. This makes the connections between atoms asymmetrical, but that is not necessarily a problem. Finally we can use the neighbors as obtained through Voronoi tessellation, i.e. two atoms are connected if they are neighboring cells in a Voronoi diagram. The connections are symmetrical and we also avoid isolated atoms. We use the software package Voro++ \citep{Rycroft2009-vt} to compute the tessellation.

\begin{figure}[tp]
	\begin{minipage}[t]{0.48\textwidth}
	\centering\includegraphics[width=1.0\textwidth]{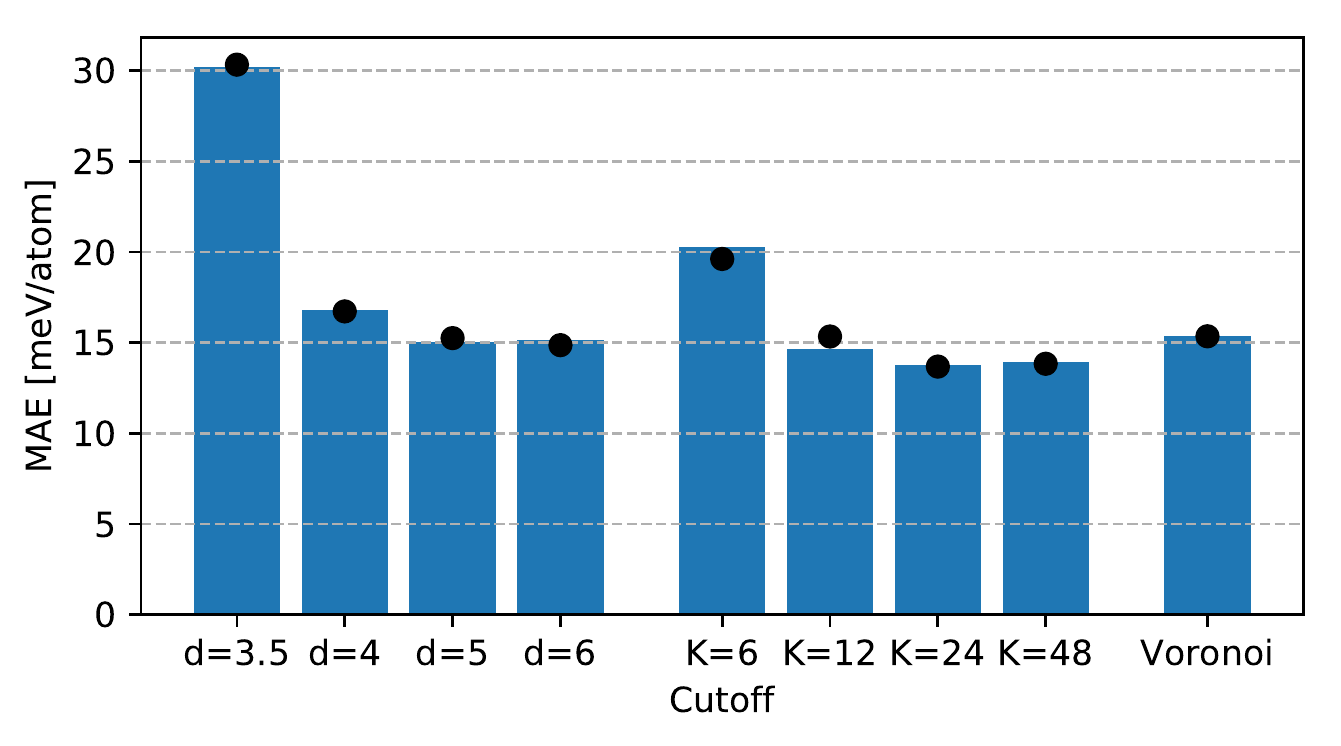}
	\captionof{figure}{MAE of formation energy predictions on OQMD using different cutoffs (cutoff distance in Ångström, K-nearest and Voronoi) for constructing the graphs. The black dots show the mean absolute error for the validation set used for early stopping. The lowest error is 13.7 meV/atom using K-nearest cutoff with K=24.}
	\label{fig:graphcuts}
	\end{minipage} \hfill
	\begin{minipage}[t]{0.48\textwidth}
	\centering\includegraphics[width=1.0\textwidth]{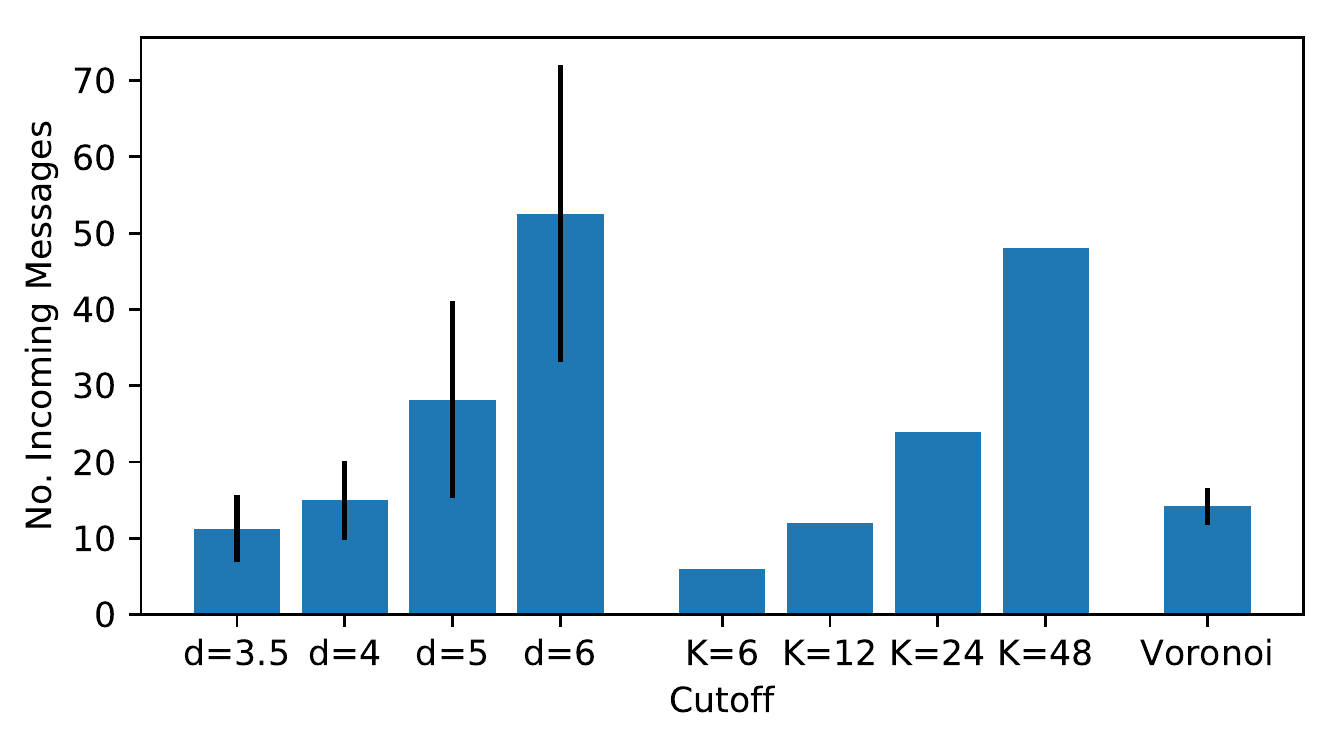}
	\captionof{figure}{Average number of incoming messages per atom for all atoms in OQMD. The error bars show the standard deviation.}
	\label{fig:edgecount}
	\end{minipage}
\end{figure}

The prediction accuracy of the model when using different cutoff methods is shown in \figref{fig:graphcuts}. We also show the average number of incoming edges across all atoms in the OQMD dataset in \figref{fig:edgecount}, which serves as a proxy for the computational complexity.
The results shows that the K-nearest cutoff is more efficient in terms of achieving a low MAE for a given average number of edges per atom and we achieve the lowest error (\SI{13.7}{meV}) with $K=24$. This is not only caused by eliminating the ``isolated atoms problem'', because the Voronoi tessellation method is also without this problem. One reason could be that training the model is more stable when the number of incoming messages is constant.

\section{Conclusion}\label{sec:conclusion}
We have proposed a novel neural message passing model for molecules and materials by extending the SchNet model
with an edge update network that allows the information exchanged between atoms to be dependent on the sending and receiving atom. This simple extension leads to a remarkable improvement in prediction accuracy.
By inspecting the learned filters for the molecular formation energy prediction task we found that the edge updates
in the first layer has a higher dependence on the receiving atom than the sending atom.
Through numerical simulations we demonstrate improvements in accuracy on formation energy prediction
of molecules and materials across all three benchmark datasets.
We also highlight the problem of choosing the cutoff when constructing the
graphs used as input for the model. We found that using K-nearest neighbors
cutoff yields lower error than using a constant cutoff distance.
We believe these results are important for applications of message
passing neural networks for predicting properties of molecules and materials and hope
to see more applications and further improvements of the model architecture in the future.






\bibliographystyle{hapalike}
\bibliography{references}


\end{document}